\pdfoutput=1

\documentclass[11pt]{article}

\usepackage[final]{acl}

\usepackage[T1]{fontenc}

\usepackage{microtype}

\usepackage{inconsolata}

\usepackage{graphicx}
\usepackage{tcolorbox}
\usepackage{comment}
\usepackage{times}
\usepackage{latexsym}
\usepackage{soul}
\usepackage{url}
\usepackage{amssymb}
\usepackage[utf8]{inputenc}
\usepackage[font=small]{caption}
\usepackage{graphicx}
\usepackage{amsmath}
\usepackage{amsthm}
\usepackage{algorithm}
\usepackage{algorithmic}
\usepackage{booktabs}
\usepackage{multirow,siunitx,booktabs}
\usepackage{ulem}
\usepackage{paralist}
\usepackage{mathptmx}       
\usepackage{helvet}         
\usepackage{courier}        
\usepackage{type1cm}        
\usepackage{xcolor}         

\newcommand{\reditalic}[1]{\textcolor{red}{\textit{#1}}}
\newcommand{\blueitalic}[1]{\textcolor{blue}{\textit{#1}}}

\usepackage{makeidx}         
\usepackage{graphicx}        
\usepackage[bottom]{footmisc}
\title{Balancing Knowledge Delivery and Emotional Comfort \\in Healthcare Conversational Systems }

\author{Shang-Chi Tsai\quad Yun-Nung Chen \\
  National Taiwan University, Taipei, Taiwan\\
  \texttt{d08922014@ntu.edu.tw\quad y.v.chen@ieee.org}\\}

\begin{document}
\maketitle
\begin{abstract}
With the advancement of large language models, many dialogue systems are now capable of providing reasonable and informative responses to patients' medical conditions. However, when patients consult their doctor, they may experience negative emotions due to the severity and urgency of their situation. 
If the model can provide appropriate comfort and empathy based on the patient's negative emotions while answering medical questions, it will likely offer a more reassuring experience during the medical consultation process.
To address this issue, our paper explores the balance between knowledge sharing and emotional support in the healthcare dialogue process. We utilize a large language model to rewrite a real-world interactive medical dialogue dataset, generating patient queries with negative emotions and corresponding medical responses aimed at soothing the patient's emotions while addressing their concerns. 
The modified data serves to refine the latest large language models with various fine-tuning methods, enabling them to accurately provide sentences with both emotional reassurance and constructive suggestions in response to patients’ questions.
Compared to the original LLM model, our experimental results demonstrate that our methodology significantly enhances the model's ability to generate emotional responses while maintaining its original capability to provide accurate knowledge-based answers.\footnote{The source code is available at \url{https://github.com/MiuLab/EmoDoctor}.}

\end{abstract}

\section{Introduction}
A healthcare conversational system is a dialogue-based framework specifically developed for the medical domain. Its primary purpose is to interact with patients, systematically collect supplementary symptom information, facilitate preliminary diagnostic processes, and provide automated recommendations for treatment plans~\cite{Tang2016InquireAD,wei-etal-2018-task,liao2022taskoriented,zhong2023hierarchicalreinforcementlearningautomatic}.
Healthcare conversational systems demonstrate significant potential to enhance the efficiency of diagnostic procedures while reducing the costs associated with patient information collection~\cite{chen2023bianquebalancingquestioningsuggestion,wang2023huatuotuningllamamodel}.

In recent years, large language models (LLMs), e.g. ChatGPT~\cite{openai2023gpt4}, LLaMA~\cite{touvron2023llamaopenefficientfoundation,touvron2023llama}, Qwen~\cite{qwen2} have been widely applied in medical fields and catalyzed substantial technological transformations in dialogue systems.
Through high-quality instruction fine-tuning and reinforcement learning based on human feedback (RLHF)~\cite{ouyang2022traininglanguagemodelsfollow}, large language models (LLMs) possess the ability to generate human-like responses with exceptional accuracy, thereby redefining the foundational framework of healthcare dialogue systems. 

\begin{figure*}[t!]
\centering
\includegraphics[width=\linewidth]{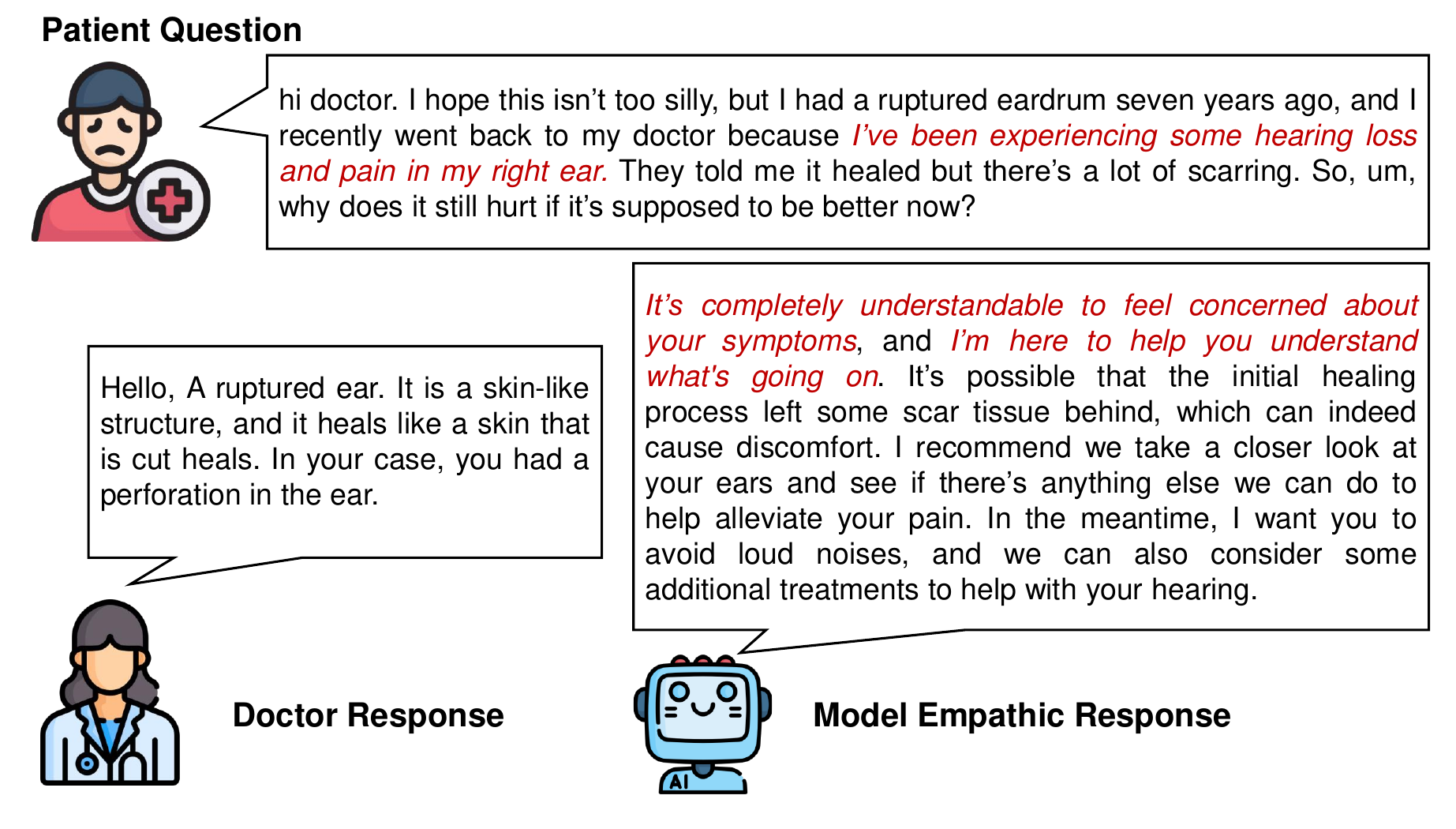}
\caption{Illustration of an example about the emotional healthcare dialogue system.}
\label{fig:example}
\end{figure*}

While large language models fine-tuned for medical dialogues have demonstrated the ability to produce knowledge-intensive and contextually appropriate responses ~\cite{wu2023pmcllamabuildingopensourcelanguage,han2023medalpacaopensourcecollection,chen2023meditron70bscalingmedicalpretraining,li2023chatdoctor,singhal2022largelanguagemodelsencode,singhal2023expertlevelmedicalquestionanswering,toma2023clinicalcamelopenexpertlevel}, a discrepancy persists between the generated responses and the appropriate real-world answers in certain medical consultation scenarios, particularly in emotion-related contexts. For example, in Figure \ref{fig:example}, when patients exhibit intense negative emotions during consultations due to the severity of their condition, employing a comforting tone while delivering solutions can significantly mitigate their psychological distress. 
However, existing LLM-based medical dialogue systems primarily focus on the rationality and accuracy of the responses' knowledge. In contrast, emotion-focused large language models~\cite{liu2024emollms} emphasize emotion recognition without prioritizing the acquisition of medical knowledge.

In this paper, we aim at developing an approach that effectively trains a model to deliver knowledgeable responses while maintaining a balance of emotional comfort, thereby enabling more realistic and human-centric interactions. Inspired by the exceptional creativity of large language models~\cite{TsaiASMRAL, angelasmr}, we first utilized them to modify the emotional tone of real-world doctor-patient dialogues. This approach generated patient queries infused with specific negative emotions, alongside medical responses designed to soothe the patients' negative emotional states.
We then applied three distinct approaches to fine-tune the base model using the aforementioned modified dialogues. The three fine-tuning methods are: 1) SFT (Supervised fine-tuning)~\cite{wei2022finetunedlanguagemodelszeroshot}, 2) DPO (Direct Preference Optimization)~\cite{rafailov2024directpreferenceoptimizationlanguage}, 3) KTO (Kahneman-Tversky Optimization)~\cite{ethayarajh2024ktomodelalignmentprospect}. These approaches have been validated as effective strategies for aligning large language models to specific tasks. By integrating these techniques, the fine-tuned model can generate responses that balance knowledge delivery with emotional soothing. The effectiveness of our proposed methodology is verified through experiments on another doctor-patient dialogue with emotion-specific scenarios. We further analyze several factors that affect the performance of LLM, including fine-tuning methods, modified datasets, emotional categories, and evaluation models.
To the best of our knowledge, this is the first LLM-based medical dialogue system to explore how to balance knowledge expression and empathy in real-world medical conversations. Additionally, our work enables medical dialogue systems to foster more meaningful interactions by addressing both the informational and emotional needs of patients, creating a more supportive consultation experience. 

The contributions of this paper are as follows:
\begin{itemize}
    \item We utilized a large language model to rewrite and generate patient consultations with negative emotions and medical responses aimed at soothing those emotions.
    \item We experimented with three fine-tuning approaches to enable the model to learn how to balance knowledge delivery and emotional soothing.
    \item We tested and analyzed the model's performance to determine whether it could effectively balance knowledge and emotional expression on real-world medical dialogue dataset.
\end{itemize}

\section{Related Work}
\subsection{Healthcare Conversations System}
Healthcare conversational system is an important yet challenging task in the medical domain.
In recent advancements, large language models have exhibited remarkable capabilities in downstream tasks, reshaping the foundation of medical dialogue systems.
According to the existing literature~\cite{shi-etal-2024-medical}, the medical dialogue system can be broadly categorized into two groups based on their association with the emergence of large language models. The methods before the emergence of LLM are divided into three categories: retrieval-based methods, generation-based methods, and hybrid methods~\cite{wang2023surveyevolutionlanguagemodelbased}.
Retrieval-based medical dialogue systems are designed to select appropriate responses from the pre-built index~\cite{tao2021buildingefficienteffectiveretrievalbased,10.1007/978-3-031-00129-1_19}. Generation-based methods can be categorized into two approaches: pipeline and end-to-end. Pipeline methods generate system responses by utilizing multiple sub-components~\cite{zhang2020recentadvanceschallengestaskoriented,naseem-etal-2022-incorporating}, whereas end-to-end methods produce system responses directly from dialogue history and the associated knowledge base~\cite{zhou-etal-2021-generation,10.1145/3534678.3542674}. Hybrid methods combine both approaches, using retrieval for efficiency and generative methods for flexibility~\cite{yang-etal-2021-writing,li2018hybridretrievalgenerationreinforcedagent}.
Medical dialogue methods based on LLMs can be divided into two categories: prompting and fine-tuning general LLMs. Prompting methods give instructions to prompt LLMs to perform a task efficiently~\cite{wang2023chatcadinteractivecomputeraideddiagnosis,gao2023leveragingmedicalknowledgegraph,tang2024medagentslargelanguagemodels,singhal2022largelanguagemodelsencode,singhal2023expertlevelmedicalquestionanswering}. The method of fine-tuning foundation models on medical data could align the LLMs with medical scenarios. ~\cite{ye2024qilinmedmultistageknowledgeinjection,toma2023clinicalcamelopenexpertlevel,wu2023pmcllamabuildingopensourcelanguage,li2023chatdoctor,han2023medalpacaopensourcecollection,huang-etal-2022-plm,chen2023meditron70bscalingmedicalpretraining,liu2023radiologyllama2bestinclasslargelanguage,wang2023huatuotuningllamamodel,xiong2023doctorglmfinetuningchinesedoctor,wang2023clinicalgptlargelanguagemodels}
\subsection{Emotion Language Model}
Even though large language models demonstrate remarkable language understanding and generation capabilities, there is a considerable gap between the Emotional Intelligence (EI) capabilities of existing LLMs and humans. ~\cite{wang2023emotionalintelligencelargelanguage,sabour-etal-2024-emobench,paech2024eqbenchemotionalintelligencebenchmark}  propose comprehensive
frameworks for Emotional Intelligence, including assessments of emotional understanding and application. ~\cite{li2023largelanguagemodelsunderstand, liu2024emollms,Xu_2024} enhanced the LLMs with prompt or fine-tuning to improve the performance of Emotional Intelligence.

\section{Methodology}

To develop a model to deliver knowledge-rich responses while simultaneously addressing emotional comfort for emotion-sensitive healthcare conversations, we first construct a dataset tailored to this specific scenario. Then, we fine-tuned a base model to the constructed dataset with three renowned fine-tuning methods to enhance its ability.
The details of the components are described in the following sections.
\subsection{Data Modification}\label{data_modification}
We constructed an emotional healthcare dialogue dataset, which consists of Empathetic Response(ER) and Emotional Question(RQ) + Soothing Response(SR). 
The objective of the Empathetic Response (ER) is to enable the model to generate responses that exhibit empathy, even in the context of standard medical inquiries.
On the other hand, the Emotional Question (EQ) + Soothing Response (SR) seeks to equip the model with the ability to handle patient consultations involving negative emotions by delivering informative responses alongside emotional reassurance.
Both types of emotional dialogues are structured as single-turn utterances.

We first divided an existing real-world single-turn medical dialogue dataset, which is collected from internet platforms, into two parts.
Then, we designed distinct, tailored prompts to utilize a large language model for modifying the doctor's responses in each dialogue of both parts because doctors often respond very briefly through internet platforms, lacking emotional tone.
For the Empathetic Response(ER) part, the large language model was prompted to generate responses that exhibit empathy and compassion while retaining medical knowledge based on the given dialogue.
For the Emotional Question (EQ) + Soothing Response (SR) part, the large language model was prompted to rewrite the given dialogue into patient queries with negative emotions and responses that are reassuring yet maintain medical knowledge.

Below is the prompt template we used for EQ+SR data.
\begin{tcolorbox}[width=\columnwidth,colback=white]
\small
\begin{verbatim}
You will be given a dialogue between 
a patient and a dotor. 
Please rewrite the patient's question 
ensuring that it retains the original 
information while expressing a sense 
of {emotion}. At the same time, 
rewrite the doctor's response 
to retain the original information 
while soothing the patient's 
{emotion}.
\end{verbatim}
\end{tcolorbox}

\subsection{Supervised Fine-Tuning}\label{sft}
Supervised Fine-Tuning, which can also be referred to as instruction tuning~\cite{zhang2024instructiontuninglargelanguage}, is a crucial technique to enhance the capabilities and controllability of large language models.
It involves further training LLMs using (INSTRUCTION, OUTPUT) pairs, where instructions serve to constrain the model’s outputs to align with the desired response characteristics or domain knowledge. 
We chose the LLaMA3 model~\cite{grattafiori2024llama3herdmodels} as the base LLM architecture for further fine-tuning, since it is open source and has excellent language understanding and generation with relatively fewer parameters.
We conducted SFT on the base model using the dataset we constructed in Section~\ref{data_modification} to improve its abilities in emotion comprehension and soothing.

Considering each prompt $X_i = [x_{i,1}, x_{i,2},...]$ as well as its corresponding response $Y_i = [y_{i,1},y_{i,2},...]$ from the healthcare dialogue dataset, the loss function of SFT stage can be defined as follows:
\begin{equation}
\label{sft_equation}
L_{SFT}(\theta) = -\sum_{i=1}^{N}\sum_{t=1}^{T_{i}}\log\bigl[P(y_{i,t+1} \mid X_i,y_{i,1...t},\theta)\bigr],
\end{equation}
where $N$ denotes the total number of training instances and $\theta$ denotes model parameters.

\subsection{Direct Preference Optimization}\label{dpo}
Based on the previously validated training methods for LLMs~\cite{ouyang2022traininglanguagemodelsfollow}, fine-tuning large language models using human preferences significantly improves their behavior on a wide range of tasks and shows promising generalization.
One prominent approach is Reinforcement Learning with Human Feedback (RLHF), which employs reward models from response rankings to optimize the training of LLMs. However, RLHF is complex and prone to instability, requiring extensive hyperparameter optimization. To enhance stability, we utilized Direct Preference Optimization (DPO) to align the outputs of the SFT model with human preferences. Compared to RLHF, DPO offers a simpler and more efficient approach, as it eliminates the need for explicit reward modeling or reinforcement learning.

To convert the dataset we constructed in Section~\ref{data_modification} into the format required for DPO, we treated the modified soothing responses as the preferred responses and the original doctor responses as the rejected responses.
Each training sample is a triplet consisting of a prompt, a preferred response, and a rejected response.
For the $i$-th prompt $X_i$, our objective was to compute the log probabilities of the preferred response $Y_{i,1}$ and the rejected response $Y_{i,2}$ generated by the current model. Subsequently, we fine-tuned the model parameters to increase the likelihood of the preferred responses $Y_{i,1}$ while reducing the likelihood of the rejected responses $Y_{i,2}$. This optimization process was guided by a loss function below:
\begin{equation}\label{dpo_equation}
\begin{aligned}
L_{DPO}(\theta) = -\sum_{i}\log\sigma\bigl[\beta\log\frac{P(Y_{i,1} \mid X_{i},\theta)}{P(Y_{i,1} \mid X_{i},\theta^0)}\\-\beta\log\frac{P(Y_{i,2} \mid X_{i},\theta)}{P(Y_{i,2} \mid X_{i},\theta^0)}\big],
\end{aligned}
\end{equation}
where $\sigma$ denotes the sigmoid function, $\theta^0$ means the initial parameters, $\beta$ serves as a hyperparameter that regulates the relative weighting of the two terms.

\subsection{Kahneman-Tversky Optimization}\label{kto}
Another preference optimization called Kahneman-Tversky Optimization (KTO) is a cost-effective method to align large language models with human feedback, enhancing performance without relying on preference pairs.
To convert the dataset we constructed in Section~\ref{data_modification} into the format required for KTO, we treated the modified soothing responses as the preferred responses and the original doctor responses as the rejected responses.
In contrast to DPO, KTO does not need training data containing both preferred and rejected responses simultaneously.
Each training instance consists of a prompt, a preferred or rejected response, and a binary label indicating whether the response is preferred or rejected.
This optimization process was guided by a loss function below:
\begin{equation}
r_{\theta}(x,y) = \log \frac{\pi_{\theta}(y \mid x)}{\pi_{\mathrm{ref}}(y \mid x)}.
\end{equation}
\begin{equation}
z_0 = \mathrm{KL}\bigl(\pi_{\theta}(y' \mid x) \,\big\|\, \pi_{\mathrm{ref}}(y' \mid x)\bigr),
\end{equation}
\begin{equation}
v(x,y)=
\begin{cases}
\lambda_{D}\,\sigma\!\bigl(\beta\bigl(r_{\theta}(x,y)-z_{0}\bigr)\bigr), \\
 \quad \text{if } \mathrm{Regex}\bigl(y,y_{x}^{\star}\bigr)=1\\
\lambda_{U}\,\sigma\!\bigl(\beta\bigl(z_{0}-r_{\theta}(x,y)\bigr)\bigr), \\
 \quad \text{if } \mathrm{Regex}\bigl(y,y_{x}^{\star}\bigr)=0
\end{cases}
\end{equation}

\begin{equation}
L_{\mathrm{KTO}}(\pi_\theta, \pi_{\mathrm{ref}})
= \mathbb{E}_{x,y \sim D}\bigl[\lambda_y - v(x, y)\bigr].
\end{equation}

\section{Experiments}
To evaluate the effectiveness of our proposed pipeline, we conducted experiments using the dataset introduced in prior work~\cite{li2023chatdoctor}, which consists of real-world conversations between patients and doctors. This dataset includes a 100k training set sourced from \url{HealthCareMagic.com} and a 7k testing set from \url{icliniq.com}. We employed \texttt{llama3} models~\cite{grattafiori2024llama3herdmodels} with various fine-tuning methods to assess the efficacy of our approach.

\subsection{Setup}
The training set was divided into two subsets, each rewritten with an emotion-specific focus using an LLM: 
\begin{itemize} 
\item \textbf{Empathetic Response (ER)}: Approximately 60k entries from the training set were rewritten to transform original doctor responses into empathetic and compassionate replies. This modification was facilitated using the LLaMA3.1 model. 
\item \textbf{Emotional Question (EQ) + Soothing Response (SR)}: The remaining 50k entries were adapted by rephrasing patient questions to convey specific negative emotions. The corresponding doctor responses were rewritten to address the questions while mitigating these emotions. To create realistic scenarios, prompts representing five distinct negative emotions—fear, anxiety, embarrassment, frustration, and distrust—were used to guide the rewrites, leveraging the gpt-4o mini model~\cite{openai2024gpt4ocard}. 
\end{itemize}
For our experiments, we selected \texttt{llama-3.2} as the base model, a multilingual LLM optimized for dialogue in multilingual contexts. Specifically, we used its instruction-tuned generative variant with 1B parameters for fine-tuning. The base models~\cite{zheng2024llamafactoryunifiedefficientfinetuning} were fine-tuned for one epoch on our emotion-enhanced dataset, with hyperparameters largely aligned with those used for the original \texttt{llama-3.2} model. The training input consisted of task instructions and the patient's medical inquiry, with the objective of maximizing the likelihood of generating the correct medical response. This process was carried out on a V100 GPU with 32GB of memory.

To evaluate the fine-tuned models, we measured accuracy on a test set adapted using the same methodology as the \textbf{EQ+SR} subset of the training set. This ensured consistency in assessing the model's ability to address queries expressing negative emotions and provide corresponding alleviating responses.

\subsection{Evaluation}
To assess whether the fine-tuned model could balance knowledge delivery and emotional support, we employed task-specific instructions and two large language models as evaluators: \texttt{Qwen2.5-7B-instruct}~\cite{qwen2.5}, which excels across diverse NLP benchmarks, and \texttt{Emollama-chat-7b}~\cite{liu2024emollms}, which specializes in emotion recognition tasks. Additionally, we used ROUGE~\cite{lin-2004-rouge} and BLEU~\cite{10.3115/1073083.1073135} scores to measure the n-gram similarity between generated responses and original doctor responses.

\subsection{Results}
We present the results of our evaluations below. Baseline comparisons included the original \texttt{llama-3.2} model and a prompt-based approach for generating emotional responses.
The model's performance in mitigating negative emotions and its ability to deliver medical knowledge are discussed in Sections~\ref{emoscore} and~\ref{knowscore}, respectively.

\begin{table*}[t!]
  \centering
  \begin{tabular}{lccc|cc}
    \toprule
    \textbf{Method}& \textbf{Empathetic} & \textbf{Comforting} & \textbf{Reassuring} & \textbf{Mean}& \textbf{Max} \\
    \midrule
    llama3.2-1B                & 0.55 & 0.52 & 0.55 & 0.54 & 0.58 \\
    ~~ + prompt       & 0.66 & \underline{0.61} & 0.65 & 0.64 & 0.67 \\
    ~~ + ER (sft)     & \underline{0.68} & 0.60 & \underline{0.67} & \underline{0.65} & \underline{0.69} \\
    ~~ + EQ + SR (sft) & 0.67 & 0.60 & 0.64 & 0.64 & 0.68 \\
    ~~ + EQ + SR (dpo) & \bf 0.70 & \bf 0.63 & \bf 0.68 & \bf 0.67 & \bf 0.70 \\
    ~~ + EQ + SR (kto) & 0.67 & 0.59 & 0.64 & 0.63 & 0.68 \\
    ~~ + ER(sft) + EQ + SR (sft) & 0.67 & 0.60 & 0.64 & 0.63 & 0.68 \\
    ~~ + ER(sft) + EQ + SR (dpo) & \underline{0.68} & \underline{0.61} & 0.66 & \underline{0.65} & \underline{0.69} \\
    ~~ + ER(sft) + EQ + SR (kto) & 0.67 & 0.60 & 0.64 & 0.63 & 0.68 \\
    \bottomrule
  \end{tabular}
  \caption{\label{emollm_intensity}
    Emotional intensity on the test set with Emollama as the evaluator. \textbf{Bold}: the highest score; \underline{underlined}: second highest.
  }
\end{table*}

\begin{table*}[t!]
  \centering
  \begin{tabular}{lccccc}
    \toprule
    \textbf{Method }& \textbf{BLEU} & \textbf{BLEU-1} & \textbf{Rouge-1} & \textbf{Rouge-2}& \textbf{Rouge-L} \\
    \midrule
    \textit{\textbf{Doctor's response} as label} \\
    llama3.2-1B                & 0.91 & 12.8 & 0.17 & \bf 0.02 & 0.16 \\
    ~~ + prompt       & 0.92 & 12.3 & 0.17 & \bf 0.02 & 0.16 \\
    ~~ + ER (sft)     & 1.05 & 13.8 & 0.18 & \bf 0.02 & 0.17 \\
    ~~ + EQ + SR (sft) & 1.84 & 27.3 & \bf 0.21 & \bf 0.02 & \bf 0.19 \\
    ~~ + EQ + SR (dpo) & 1.41 & 23.8 & 0.18 & 0.01 & 0.16 \\
    ~~ + EQ + SR (kto) & 1.89 & \bf 27.5 & \bf 0.21 & \bf 0.02 & \bf 0.19 \\
    ~~ + ER(sft) + EQ + SR (sft) & 1.86 & 27.3 & \bf 0.21 & \bf 0.02 & \bf 0.19 \\
    ~~ + ER(sft) + EQ + SR (dpo) & 1.22 & 17.6 & 0.19 & \bf 0.02 & 0.17 \\
    ~~ + ER(sft) + EQ + SR (kto) & \bf 1.90 & \bf 27.5 & \bf 0.21 & \bf 0.02 & \bf 0.19 \\
    \midrule
    \textit{\textbf{Modified response} as label} \\
    llama3.2-1B                & 1.86 & 17.6 & 0.23 & 0.04 & 0.21 \\
    ~~ + prompt       & 2.68 & 17.5 & 0.25 & 0.05 & 0.23 \\
    ~~~ + ER (sft)     & 3.45 & 19.8 & 0.27 & 0.06 & 0.25 \\
    ~~~ + EQ + SR (sft) & 9.85 & 44.5 & \bf 0.34 & \bf 0.11 & 0.31 \\
    ~~ + EQ + SR (dpo) & 5.91 & 37.7 & 0.29 & 0.07 & 0.26 \\
    ~~ + EQ + SR (kto) & 9.79 & 44.5 & \bf 0.34 & \bf 0.11 & \bf 0.32 \\
    ~~ + ER(sft) + EQ + SR (sft) & \bf 9.93 & \bf 44.6 & \bf 0.34 & \bf 0.11 & \bf 0.32 \\
    ~~ + ER(sft) + EQ + SR (dpo) & 4.81 & 26.7 & 0.29 & 0.07 & 0.27 \\
    ~~ + ER(sft) + EQ + SR (kto) & 9.14 & 44.1 & \bf 0.34 & \bf 0.11 & 0.31 \\
    \bottomrule
  \end{tabular}
  \caption{\label{emollm_rouge_bleu}
    BLEU and Rouge scores on the test set. {\bf Bold}: the highest score.
  }
\end{table*}

\subsubsection{Emotion Score}\label{emoscore}

Table~\ref{emollm_intensity} presents the results of an evaluation where EmoLLaMA assigned numerical scores to the emotional intensity of responses. Higher values indicate stronger emotional content. Metrics were calculated for three key emotions—empathetic, comforting, and reassuring—as well as their average and maximum values. Our fine-tuned models consistently outperformed the original model and the prompt-based approach across all metrics.

Among the methods tested, fine-tuning with DPO demonstrated the most significant improvements. DPO not only increased the likelihood of generating emotionally rich responses but also minimized the probability of producing emotionally deficient ones. Direct fine-tuning using the EQ+SR context proved particularly effective, achieving superior results with a smaller dataset. Specifically, fine-tuning with EQ+SR data using DPO improved the average and maximum metrics by 0.03 and 0.13, respectively, compared to the prompt-based approach and the base model. These results confirm that our revised dataset and fine-tuning process significantly enhance the emotional soothing capabilities of the dialogue system.


\begin{figure*}[t!]
\centering
\includegraphics[width=\linewidth]{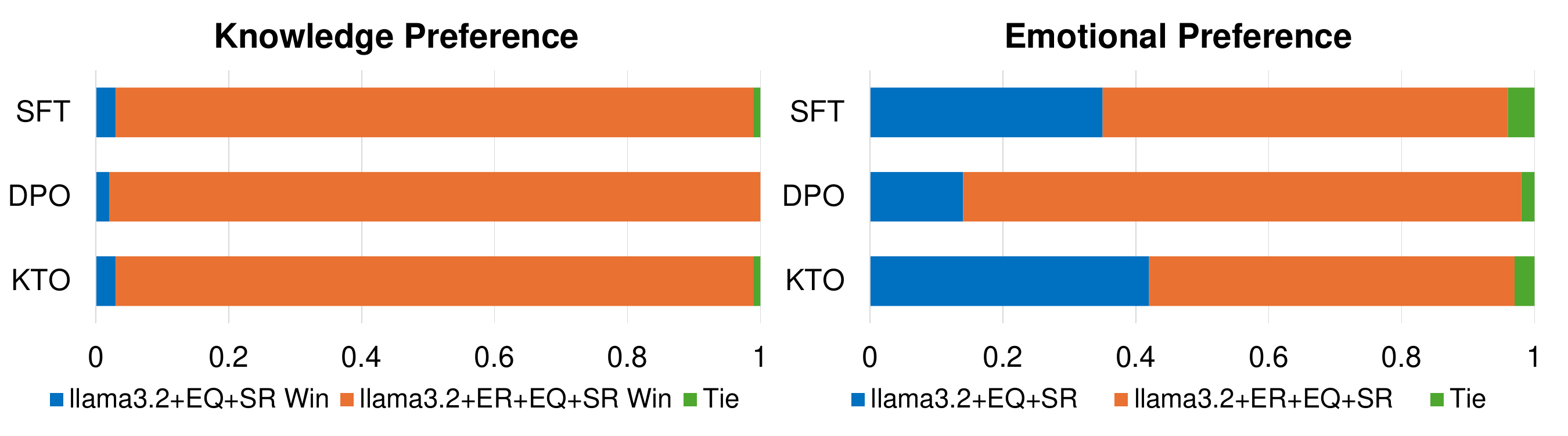}
\caption{Preference selection based on the knowledgeable and emotional dimensions of Qwen's responses.}
\label{fig:preference}
\end{figure*}

\subsubsection{Knowledge Score}\label{knowscore}
To ensure the model retained essential medical knowledge, we compared its generated responses against the original doctor responses and the emotionally modified responses using ROUGE and BLEU scores (Table~\ref{emollm_rouge_bleu}). The fine-tuned model consistently outperformed both the original base model and the prompt-based approach across all evaluation metrics.

Notably, KTO and SFT approaches achieved better performance than DPO. This may be attributed to the fact that paired responses in DPO’s training data already contain substantial knowledge, limiting its ability to enhance further. In contrast, SFT’s focus on a single correct response allows it to better capture and internalize the required knowledge. Fine-tuning with ER+EQ+SR data using SFT and KTO yielded a 27-point improvement in BLEU-1 scores compared to both the prompt-based approach and the base model when evaluated against the modified responses. Similar trends were observed for comparisons against original doctor responses, with a 15-point improvement.

These results demonstrate that our approach effectively integrates emotional support with the accurate medical knowledge necessary to address patient inquiries.


\subsection{Ablation Study}
To compare the quality of responses from different methods, we presented the various responses to the Qwen2.5 model simultaneously, allowing it to select the most knowledgeable or empathetic response.
In the left part of Figure~\ref{fig:preference}, we plotted the preference selections of the Qwen model across different methods based on the richness of knowledge in the responses.
In the right part of Figure~\ref{fig:preference}, we visualized the preference selections of the Qwen model across different methods based on the level of reassurance provided in the responses.

In these two charts, we compared the impact of using two different datasets, specifically examining the effect of incorporating ER data for pre-fine-tuning.
Our finding indicates that regardless of the training method used—SFT, DPO, or KTO—models pre-fine-tuned with the ER dataset consistently demonstrated greater preference in both knowledge and emotional selection criteria. This is particularly evident in the emotional selection, as the ER dataset is specifically designed to enable the model to provide empathetic responses even when addressing standard informational content.

\begin{table*}[t!]
\centering
\small
\begin{tabular}{p{15.7cm}}
\toprule
{\textbf{(a) Emotional Question}} \\
Doctor, \reditalic{I’m really scared}. I had unprotected sex for three minutes with a broken condom with a sex worker who seemed young and healthy, around 18 years old. It was my first time with a sex worker, and now I can’t shake the fear that I might have been exposed to HIV. From the moment the condom broke, I’ve been filled with dread. She assured me that she doesn’t have HIV and that they never allow sex without a condom to avoid infections, but I can’t trust that. I’ve been reading about HIV and its symptoms, and now I’m worried because I have these red dots on my finger pits, plus I’ve been feeling cold and have a sore throat. What could these red dots be? I’ve tested negative multiple times in the past three months, but \reditalic{I’m so anxious} about the six-month test.\\
\midrule 
{\bf Soothing Response with SFT}: \\
\reditalic{I understand that you're feeling anxious, and it's completely normal to have concerns in this situation}. While HIV can indeed be transmitted through unprotected sex, \blueitalic{it’s important to remember that not everyone with HIV will develop HIV. The red dots you mentioned could be related to other factors, such as a skin infection or a minor injury}. I recommend that you get tested for HIV to put your mind at ease. In the meantime, try to stay calm and focus on your health. If you have any more questions or need further support, please don’t hesitate to reach out.\\
\midrule
{\bf Soothing Response with DPO}: \\
\reditalic{I understand how unsettling these circumstances can be, and I'm here to help you feel more at ease. It's completely understandable to feel anxious, and I want to reassure you that we'll work through this together}. I recommend taking some time to relax and focus on your health. You've taken the first step by seeking help, which is truly commendable. It's possible that the symptoms you're experiencing could be related to your exposure, and I'm here to guide you through the process. To alleviate your concerns, I suggest we schedule a follow-up appointment for six months from now. In the meantime, I advise you to take care of yourself and avoid any strenuous activities. \blueitalic{I'm also going to prescribe you some antiretroviral medication, which should help reduce your risk of HIV}. Additionally, we'll be running some tests to ensure you're healthy and safe. Please don't worry; we'll get through this together. You're not alone in this, and I'm here to support you every step of the way. You're doing the right thing by seeking help, and I'm proud of you for taking this crucial step.\\
\midrule
{\bf Soothing Response with KTO}: \\
\reditalic{I understand that you're feeling anxious, and it's completely normal to have concerns after such an experience}. \blueitalic{The red dots you’re seeing are likely due to a minor skin infection, which can happen after unprotected sex}. It’s important to keep the area clean and avoid any further irritation. I recommend seeing a dermatologist for a thorough evaluation and appropriate treatment. Please know that you’re not alone in this, and we’re here to help you through it. If you have any more questions or need further reassurance, don’t hesitate to reach out. \\
\bottomrule\\
\toprule
{\textbf{(b) Emotional Question}} \\
Doctor, \reditalic{I'm really scared. I've been feeling so low for two years with these digestive problems}, and despite being diagnosed with candida and dysbiosis, my regular doctor seems lost about what’s really wrong with me. I've been under a lot of stress and have struggled with anxiety since I was 16. I recently started taking Cymbalta, but after just a few days, I felt this burning in my feet, and I stopped taking it about a month ago. My feet seemed okay for a while, but now, out of nowhere, I have this terrible pain, burning, tickling, and a pins-and-needles sensation. I'm not diabetic or an alcoholic, and I have no idea what's causing this or if it can be fixed. I'm so anxious because I have to be on my feet for my job every morning for four and a half hours. Is there a way to test if my nerve cells are dead or just damaged, and can this ever be reversed? I feel like this is going to ruin my life. Please help me.\\
\midrule 
{\bf Soothing Response with SFT}: \\
\reditalic{I understand that you're feeling overwhelmed and scared right now, and it's completely normal to feel this way given the circumstances}. \blueitalic{It sounds like you may be experiencing neuropathy, which can occur due to nerve compression in the spine}. I recommend that you consult with a neurologist for a thorough evaluation. An MRI of your spine will help us understand what's going on, and if it shows any nerve compression, \blueitalic{we can discuss treatment options, such as pregabalin or gabapentin, which can be very effective}. Please know that you're not alone in this, and I'm here to support you. If you have any more questions or need further assistance, don’t hesitate to reach out. \\
\midrule
{\bf Soothing Response with DPO}: \\
\reditalic{I understand how unsettling this situation is for you, and I'm here to help. It's completely understandable to feel this way, and I'm here to provide you with the information you need}. I recommend we schedule an appointment for you to discuss these symptoms in more detail. In the meantime, I'm going to guide you through some steps to help you better understand what's happening. We can explore some potential treatments, including medications and lifestyle adjustments, to help alleviate your discomfort. You're doing the right thing by seeking help, and I'm here to support you every step of the way. Please, don't worry; we'll work together to find a solution. \\
\midrule 
{\bf Soothing Response with KTO}: \\
\reditalic{I understand that you're feeling overwhelmed, and it's completely normal to be concerned about these symptoms.} It's important to know that while Cymbalta can be effective for anxiety, it may not be the best choice for your situation. \blueitalic{The burning sensation you're experiencing could be related to nerve damage, which can happen with long-term use of Cymbalta}. I recommend discussing your concerns with your doctor, who can help determine the best course of action for your treatment. Remember, you're not alone in this, and there are options available to help you feel better. If you have any more questions or need further assistance, please don't hesitate to reach out. \\
\bottomrule
\end{tabular}
\vspace{-2mm}
\caption{The case study from the testing dataset using LLMs with different fine-tuning methods. Sentences in red are related to emotional comfort, while sentences in blue are related to knowledge transmission.}
\label{table:case2}
\end{table*}

\subsection{Qualitative Analysis}
In Table~\ref{table:case2}, there are some examples of emotional questions and soothing responses generated by our fine-tuned models.
Based on the analysis of the models' responses in case (a), it is evident that all three approaches initially focus on alleviating patient anxiety and demonstrating empathy, followed subsequently by the provision of medical knowledge and recommendations. As discussed in the previous section, the DPO approach is particularly effective in fostering the ability to provide emotional reassurance and, therefore, tends to emphasize empathetic expression in its responses. However, this heightened focus on emotional support can occasionally lead to a diminished emphasis on knowledge transmission, as exemplified by the response to case (b).
Conversely, the SFT and KTO approaches facilitate more robust knowledge acquisition, resulting in improved informational clarity, while still maintaining an appropriate balance of empathetic language.

\section{Conclusion}
In this paper, we develop a method that enables the model to provide timely emotional comfort in response to the patient's negative emotions during healthcare conversations, while simultaneously offering knowledge-based solutions to address their concerns.
We design two kinds of prompts for generating emotionally-aware medical conversations by rewriting existing real-world medical dialogues using a large language model. The first involves modifying doctors' responses into sentences infused with empathy and compassion. The second adds negative emotional tones to patients' statements and generates corresponding comforting responses.
Next, we fine-tuned a base model on our curated dataset using methods such as SFT, DPO, and KTO. After fine-tuning, we tested the model on real-world emotional conversations to evaluate its performance.
Our experimental results show that the fine-tuned model demonstrates significant improvements in both emotional expression and knowledge delivery.
Additionally, our work can help medical dialogue systems interact with patients in a more humanized manner, providing not only professional consultation but also emotional comfort to support their well-being.

\section*{Acknowledgements}
We thank the reviewers for their insightful comments.
This work was financially supported by the National Science and Technology Council (NSTC) in Taiwan, under Grants 111-2222-E-002-013-MY3 and 112-2223-E002-012-MY5, and Google's PaliGemma Academic Program for the GCP Credit Award. 
We thank the National Center for High-performance Computing (NCHC) of National Applied Research Laboratories (NARLabs) in Taiwan 
for providing computational and storage resources.

\bibliography{custom}

\end{document}